\documentclass[twoside,11pt]{article}

\usepackage{jmlr2e}
\usepackage{svg}

\usepackage{subfigure}
\usepackage[utf8]{inputenc} 
\usepackage[T1]{fontenc}    
\usepackage{hyperref}       
\usepackage{url}            
\usepackage{booktabs}       
\usepackage{amsfonts}       
\usepackage{nicefrac}       
\usepackage{microtype}      
\usepackage{xcolor}         
\usepackage[normalem]{ulem}
\usepackage[acronym]{glossaries}
\usepackage{xspace}

\usepackage{graphicx}
\usepackage{pythonhighlight}
\usepackage{bold-extra}
\usepackage{enumitem}

\usepackage{listings}
\usepackage{color}
\usepackage{lastpage} 

\definecolor{dkgreen}{rgb}{0,0.6,0}
\definecolor{gray}{rgb}{0.5,0.5,0.5}
\definecolor{mauve}{rgb}{0.58,0,0.82}

\newcommand{\eg}{{e.g.}\xspace}

\newacronym{ai}{AI}{artificial intelligence}
\newacronym{ml}{ML}{machine learning}
\newacronym{XAI}{XAI}{eXplainable artificial intelligence}

\usepackage{lastpage}
\jmlrheading{24}{2023}{1-\pageref{LastPage}}{2/22; Revised
11/22}{1/23}{22-0142}{Anna Hedström, Leander Weber, Dilyara Bareeva, Daniel Krakowczyk, Franz Motzkus, Wojciech Samek, Sebastian Lapuschkin, and Marina M.-C. Höhne}
\ShortHeadings{Quantus: An XAI Toolkit for Evaluating Explanations}{Hedström, Weber, Bareeva, Krakowczyk, Motzkus, Samek, Lapuschkin, and Höhne}

\begin{document}

\title{Quantus: An Explainable AI Toolkit for Responsible \\ Evaluation of Neural Network Explanations and Beyond} 

\author{
\name Anna Hedström$^{1,\dagger}$
\email anna.hedstroem@tu-berlin.de
\AND
\name Leander Weber$^3$
\email leander.weber@hhi.fraunhofer.de
\AND
\name Dilyara Bareeva$^1$
\email dilyara.bareeva@campus.tu-berlin.de
\AND
\name Daniel Krakowczyk$^4$
\email daniel.krakowczyk@uni-potsdam.de
\AND
\name Franz Motzkus$^3$
\email franz.motzkus@hhi.fraunhofer.de
\AND
\name Wojciech Samek$^{2,3,5}$
\email wojciech.samek@hhi.fraunhofer.de
\AND
\name Sebastian Lapuschkin$^{3,\dagger}$
\email sebastian.lapuschkin@hhi.fraunhofer.de
\AND
\name Marina M.-C. Höhne$^{1,5,\dagger}$
\email marina.hoehne@tu-berlin.de\\
\addr
$^1$ Understandable Machine Intelligence Lab, 
TU Berlin,
10587 Berlin,
Germany\\
$^2$ Department of Electrical Engineering and Computer Science, TU Berlin, 10587 Berlin, Germany\\
$^3$ Department of Artificial Intelligence,
Fraunhofer Heinrich-Hertz-Institute,
10587 Berlin,
Germany\\
$^4$ Department of Computer Science,
University of Potsdam,
14476 Potsdam,
Germany\\
$^5$ BIFOLD – Berlin Institute for the Foundations of Learning and Data,
10587 Berlin,
Germany\\
$^\dagger$ corresponding authors
}

\editor{Joaquin Vanschoren} 

\maketitle

\begin{abstract}
The evaluation of explanation methods is a research topic that has not yet been explored deeply, 
however, since explainability is supposed to strengthen trust in artificial intelligence, it is necessary to systematically review and compare explanation methods 
in order to confirm their correctness.
Until now, no tool with focus on XAI evaluation exists that exhaustively and speedily allows researchers to evaluate the performance of explanations of neural network predictions.
To increase transparency 
and reproducibility in the field,
we therefore built \texttt{Quantus}---a comprehensive, evaluation toolkit in Python that includes a growing, well-organised collection of evaluation metrics and tutorials 
for evaluating explainable methods.
The toolkit has been thoroughly tested and is available under an
open-source license on PyPi (or on \url{https://github.com/understandable-machine-intelligence-lab/Quantus/}).
\end{abstract}


\section{Introduction}

Despite much excitement and activity in the field of \gls{XAI} \citep{montavon2018methods, arya2019explanation, LapNCOMM19,  SamPIEEE21, bykov2021explaining}, 
the evaluation of explainable methods still remains an unsolved problem 
\citep{samek2016evaluating, adebayo2020debugging, 2020scs, yona2021revisiting, arras2021ground}. 
Unlike in traditional \gls{ml}, 
the task of \emph{explaining} generally lacks ``ground-truth'' data. 
There exists no universally accepted definition of what a ``correct’' explanation is, or what properties an explanation should fulfil \citep{BAM2019}.
Due to this lack of standardised evaluation procedures in \gls{XAI},
researchers frequently conceive new ways to experimentally examine explanation methods
\citep{bach2015pixel, samek2016evaluating, adebayo2020sanity,
BAM2019,
kindermans2019reliability},
oftentimes employing different parameterisations and
various kinds of preprocessing and normalisations,
each leading to different or even contrasting results,
making evaluation outcomes
difficult to interpret and compare.
Critically, we note that it is common for \gls{XAI} papers to base their conclusions on one-sided, sometimes methodologically questionable evaluation procedures, which we fear may hinder access to the current State-of-the-art (SOTA) in \gls{XAI} and potentially hurt the perceived credibility of the field over time. 

For these reasons, 
researchers often rely on a qualitative evaluation of explanation methods (\eg, \cite{zeiler2014visualizing, ribeiro2016why, shrikumar2017learning}).
Although qualitative evaluation of XAI methods is an important and complementary type of evaluation analysis \citep{hoffman},
the assumption that humans are able to recognise a correct explanation comes with a series of pitfalls:
not only does the notion of an ``accurate'' explanation often depend on the specifics of the task at hand,
humans are also questionable judges of quality \citep{wang2019designing, rosenfeld2021better}. 
In addition, 
recent studies suggest that even 
quantitative evaluation 
of explainable methods
is far from fault-proof \citep{bansal2020sam, budding2021evaluating, yona2021revisiting, hase2021the}. 
In response to these issues, 
we developed \texttt{Quantus},  to provide the community with a 
versatile and comprehensive toolkit
that collects, organises, and explains
a wide range of evaluation metrics 
proposed for explanation methods.
The library is designed to help
automate the process of \emph{XAI quantification}---by
delivering speedy, easily digestible, and
at the same time holistic summaries 
of the quality of the given explanations. 
As we see it,
\texttt{Quantus} concludes 
an important, still missing contribution
in today's \gls{XAI} research by
filling the gap between what the community produces and what it currently needs:
a more quantitative, 
systematic
and standardised
evaluation of explanation methods.

\section{Toolkit Overview}

\texttt{Quantus} provides its intended users---practitioners and researchers interested in the domains of \gls{ml} and \gls{XAI}---with a steadily expanding list of 30+ reference metrics
to evaluate explanations of \gls{ml} predictions. Moreover, it offers
comprehensive guidance on how to use these metrics, including information about potential pitfalls in their application.

\vspace{-0.75em} 
\begin{table*}[h!]

  \caption{\small{Comparison of four \gls{XAI} libraries---(\texttt{AIX360} \citep{arya2019explanation}, \texttt{captum} \citep{kokhlikyan2020captum}, \texttt{TorchRay} \citep{fong2019understanding} and \texttt{Quantus}) in terms of the number of \gls{XAI} evaluation methods for six different evaluation categories, as implemented in each library.
  }}
  \label{comparison-table}
  \centering
  \footnotesize
  \begin{tabular}{l|cccccc}
    \toprule
    Library & Faithfulness
    & Robustness & Localisation & Complexity & Axiomatic  & Randomisation \\
    \midrule
    \texttt{Captum} (2) & 1 & 1 & 0 & 0 & 0 & 0 \\ 
    \texttt{AIX360} (2) & 2 & 0 & 0 & 0 & 0 & 0  \\ 
    \texttt{TorchRay} (1) & 0 & 0 & 1 & 0 & 0 & 0 \\ 
    \texttt{\textbf{Quantus}} (27) & \textbf{9} & \textbf{4} & \textbf{6} & \textbf{3} & \textbf{3} & \textbf{2} \\
    \bottomrule
  \end{tabular}
\end{table*}

\begin{figure}[!t]
    \centering
    \includegraphics[width=0.75\textwidth]{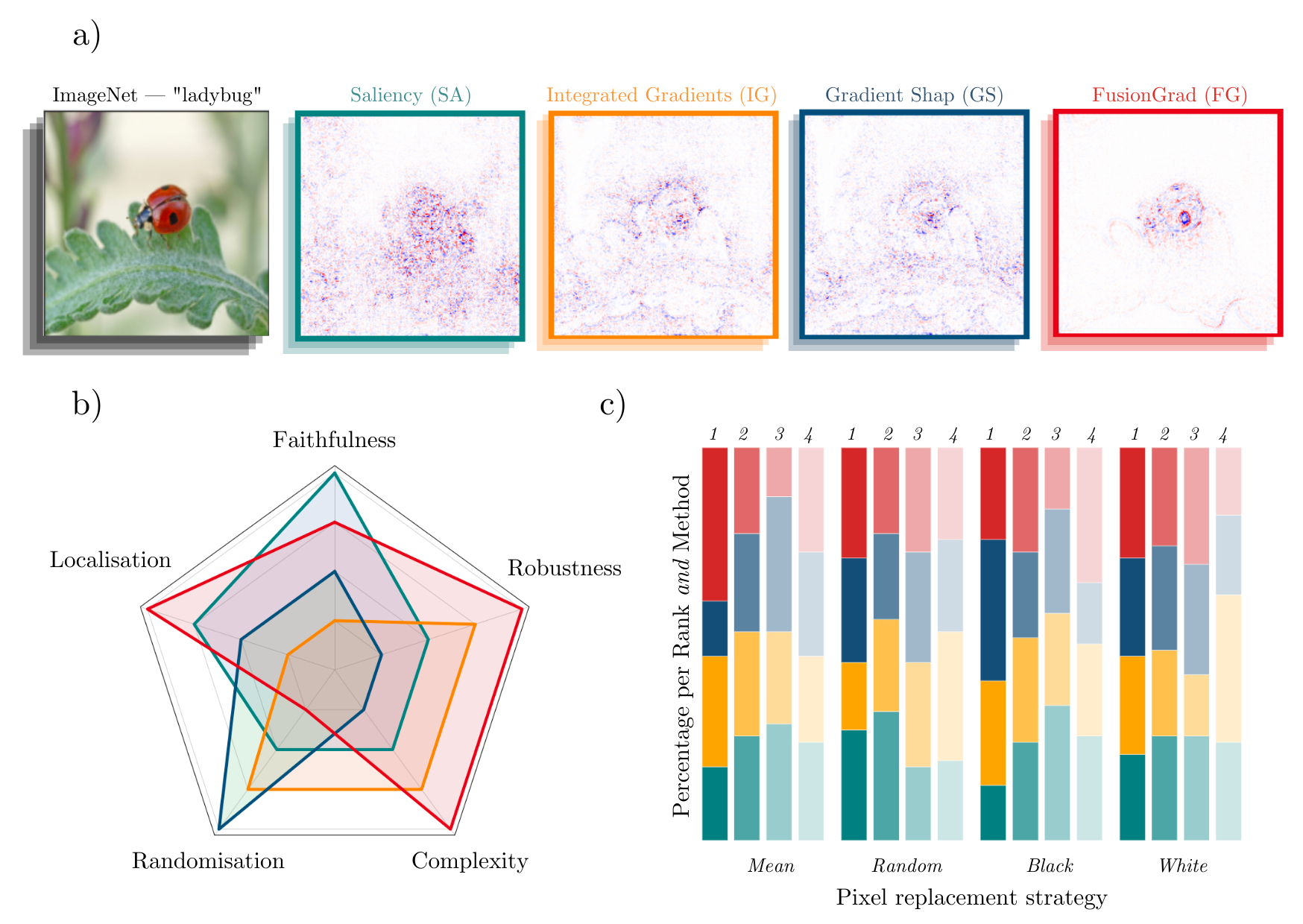}
    \caption{\small{\emph{a)} Simple \textit{qualitative} comparison of XAI methods is often not sufficient to distinguish which gradient-based method---Saliency \citep{morch, baehrens}, Integrated Gradients \citep{sundararajan2017axiomatic}, GradientShap \citep{lundberg2017unified} or FusionGrad \citep{bykov2021noisegrad} is preferred. With \texttt{Quantus}, we can obtain richer insights on how the methods compare \emph{b)} by holistic quantification on several evaluation criteria and \emph{c)} by providing sensitivity analysis of how a single parameter, \eg, pixel replacement strategy of a faithfulness test influences the ranking of explanation methods. 
    }}
    \label{image}
    \vspace{-1.5em}
\end{figure}

The library is thoroughly documented and
includes tutorials covering multiple use-cases, data domains
and tasks---from comparative analysis of \gls{XAI} methods and attributions, to quantifying the extent evaluation outcomes are dependent on metrics' parameterisations. In Figure \ref{image}, 
we demonstrate
some example analysis using ImageNet dataset \citep{ILSVRC15}
that can be produced 
with \texttt{Quantus}\footnote{The full experiment can be reproduced (and obtained) at the repository, under the \texttt{\textbackslash tutorials} folder.}.
The library provides an abstract layer between APIs of deep learning frameworks, \eg, \texttt{PyTorch} \citep{NEURIPS2019_9015} and \texttt{tensorflow} \citep{tensorflow2015-whitepaper} and
can be employed iteratively both during and after model training.
Code quality is ensured by thorough testing, using \texttt{pytest} and continuous integration (CI), where every new contribution is
automatically checked for sufficient test coverage. We employ 
syntax formatting 
with \texttt{flake8}, \texttt{mypy} and \texttt{black}
under various Python versions.

Unlike other \gls{XAI}-related libraries\footnote{
Related libraries were selected with respect to the \gls{XAI} evaluation capabilities. Packages including no metrics for evaluation of explanation methods,
\eg, \texttt{Alibi} \citep{alibijanis}, \texttt{iNNvestigate} \citep{innvestigatealber},
\texttt{dalex} \citep{dalexhubert} 
and \texttt{zennit} \citep{anders2021software} were excluded.},
\texttt{Quantus} has its primary focus on evaluation and
as such, 
supports a breadth of metrics,
spanning various evaluation categories (see \autoref{comparison-table}).
 A detailed description
of the different evaluation categories
can be found in
the Appendix.
The first iterations of the library
mainly focus on
attribution-based explanation techniques\footnote{This category of explainable methods aims to assign an importance value to the model features and arguably, 
is the most studied group of explanations.} 
for (but not limited to) image classification.  
In planned future releases, 
we are working towards 
extending the applicability of the library further,
\eg, 
by 
developing additional metrics and 
functionality
that will enable users
to perform checks, 
verifications and sensitivity analyses on top of the metrics.

\section{Library Design}

The user-facing API of \texttt{Quantus} is designed 
with the aim of replacing an oftentimes lengthy and open-ended evaluation procedure
with structure and speed---with a single line of code, 
the user can gain quantitative insights
of how their explanations are behaving under various criteria. 
In the following code snippet, 
we demonstrate 
one way for
how \texttt{Quantus} can be used
to evaluate pre-computed explanations via a
\texttt{PixelFlipping} experiment \citep{bach2015pixel}. 
In this example, we assume 
to have 
a pre-trained model (\texttt{model}), 
a batch of input and output pairs (\texttt{x\_batch}, \texttt{y\_batch}) 
and
a set of attributions
(\texttt{a\_batch}).
\vspace{0.5em}
\begin{lstlisting}[style=mypython, label=code-snippet]
import quantus
pixelflipping = quantus.PixelFlipping(perturb_baseline="black", abs=True)
scores = pixelflipping(model, x_batch, y_batch, a_batch, **params)
pixelflipping.plot(y_batch=y_batch, scores=scores)
\end{lstlisting}
\vspace{0.5em}

Needless to say, \gls{XAI} evaluation is intrinsically difficult and
there is no one-size-fits-all metric for all tasks. Evaluation of explanations must, therefore, be understood and calibrated from its context:
the application, 
data, 
model,
and intended stakeholders \citep{chander2018evaluating,arras2021ground}. 
To this end, we designed \texttt{Quantus} to be highly customisable and easily extendable---API documentation and examples on how to create new metrics
as well as how to customise existing ones
are included.  
Thanks to the API, 
any supporting functions of the evaluation procedure, \eg, 
\texttt{perturb\_baseline} that determines the value that the input features should be iteratively masked with,
can flexibly be replaced by a user-specified function
to ensure that the evaluation procedure is appropriately contextualised.  

It is practically well-known
but not yet publicly recognised
that evaluation outcomes of explanations
can be highly sensitive to the parameterisation of metrics \citep{bansal2020sam, agarwal2020explaining} and other confounding factors introduced in the evaluation procedure \citep{hase2021out, yona2021revisiting}. 
Therefore,
to encourage a thoughtful and responsible selection and parameterisation of metrics,
we added mechanisms such as warnings,
checks
and user guidelines,
cautioning users to reflect upon their choices.

\section{Broader Impact}

We built \texttt{Quantus} to
raise the bar of \textit{\gls{XAI} quantification}---to substitute an ad-hoc and sometimes ineffective evaluation procedure
with reproducibility, simplicity and transparency.
From our perspective, 
\texttt{Quantus} contributes to the \gls{XAI} development
by helping researchers to
speed up the development and application of explanation methods,
dissolve existing ambiguities and
enable more comparability. 
As we see it, steering efforts towards increasing
objectiveness of evaluations and reproducibility in the field will prove rewarding for the community as a whole.
We are convinced that a holistic, 
multidimensional take on \gls{XAI} quantification will be imperative to the general success of (X)AI over time. 

\acks{This work was partly funded by the German Federal Ministry for Education and Research through project Explaining 4.0 (ref. 01IS20055), BIFOLD (ref. 01IS18025A and ref. 01IS18037A), AEye (ref. 01IS20043),
the Investitionsbank Berlin through BerDiBA (grant no. 10174498),
as well as the European Union’s Horizon 2020 programme through iToBoS (grant no. 965221).}

\section*{Appendix}

In most explainability contexts, ground-truth explanations are not available \citep{samek2016evaluating, adebayo2020debugging, 2020scs, yona2021revisiting, arras2021ground}, which makes the task of evaluating explanations non-trivial. Efforts on evaluating explanations have therefore been invested diversely. For better organisation, in the source code of \texttt{Quantus}, we therefore grouped the metrics into six categories based on their logical similarity---(a) faithfulness, (b) robustness, (c) localisation, (d) complexity, (e) randomisation and (f) axiomatic metrics. 

In the following, we describe each of the categories briefly. A more in-depth description of each category, including an account of the underlying metrics, is documented in the repository. The direction of the arrow indicates whether higher or lower values are considered better (exceptions within each category exist, so please carefully read the docstrings of each individual metric prior to usage and/or interpretation).

\begin{enumerate}[label=(\alph*)]
    \item \textit{Faithfulness} ($\uparrow$) quantifies to what extent explanations follow the predictive behaviour of the model, asserting that more important features affect model decisions more strongly \citep{bhatt2020, alvarezmelis2018robust, arya2019explanation, nguyen2020, bach2015pixel, samek2016evaluating, montavon2018methods, ancona2019, irof2020, yeh2019, rong2022, dasgupta2022}
    \item \textit{Robustness} ($\downarrow$) measures to what extent explanations are stable when subject to slight perturbations in the input, assuming that the model output approximately stayed the same \citep{yeh2019, montavon2018methods, alvarezmelis2018robust, dasgupta2022} 
    \item \textit{Localisation} ($\uparrow$)  tests if the explainable evidence is centred around a region of interest, which may be defined around an object by a bounding box, a segmentation mask or a cell within a grid \citep{pg2018, theiner2021, kohlbrenner2020best, arras2021ground, rong2022, focus2022}
    \item \textit{Complexity} ($\downarrow$) captures to what extent explanations are concise, i.e., that few features are used to explain a model prediction \citep{chalasani2020, bhatt2020, nguyen2020}
    \item \textit{Randomisation} ($\uparrow$) tests to what extent explanations deteriorate as the data labels or the model, e.g., its parameters are increasingly randomised \citep{adebayo2020sanity, sixt2019} 
    \item \textit{Axiomatic} ($\uparrow$) measures if explanations fulfill certain axiomatic properties \citep{kindermans2019reliability, sundararajan2017axiomatic, nguyen2020} 
\end{enumerate}

\bibliography{refs.bib}

\end{document}